\documentclass[letterpaper, 10 pt, conference]{ieeeconf}  % Comment this line out if you need a4paper

\IEEEoverridecommandlockouts                              % This command is only needed if 
\usepackage{multicol}
\usepackage{algorithm}
\usepackage{algorithmicx}
\usepackage[noend]{algpseudocode}
\usepackage{amsmath}
\usepackage{booktabs}
\usepackage{graphicx}
\usepackage{amssymb} 
\usepackage{dsfont}
\usepackage{subfig}
\usepackage{wrapfig}
\usepackage[font=small,labelfont=bf]{caption}
\usepackage{xcolor}
\usepackage{graphbox}
\usepackage{hyperref}
\usepackage{balance}

\DeclareMathOperator*{\argmax}{arg\,max}
\DeclareMathOperator*{\argmin}{arg\,min}

\graphicspath{ {./data/} }

\title{}
\title{\LARGE \bf
Synergistic Scheduling of Learning and Allocation of Tasks in Human-Robot Teams
}

\author{Shivam Vats$^{1}$ \and Oliver Kroemer$^{1}$ \and  Maxim Likhachev$^{1}$% <-this % stops a space
\thanks{$^{1}$Robotics Institute, Carnegie Mellon University
\tt \small \{svats, okroemer, mlikhach\} @andrew.cmu.edu
}
}

\begin{document}
\maketitle
\thispagestyle{empty}
\pagestyle{empty}

\begin{abstract}
We consider the problem of completing a set of $n$ tasks with a human-robot team using minimum effort. 
In  many domains, teaching a robot to be fully autonomous can be counterproductive if there are finitely many tasks to be done.
Rather, the optimal strategy is to weigh the \textit{cost} of teaching a robot and its \textit{benefit}- how many new tasks it allows the robot to solve autonomously.
We formulate this as a planning problem where the goal is to decide what tasks the robot should do autonomously (act), what tasks should be delegated to a human (delegate) and what tasks the robot should be taught (learn) so as to complete all the given tasks with minimum effort.
This planning problem results in a search tree that grows exponentially with $n$ -- making standard graph search algorithms intractable.
We address this by converting the problem into a mixed integer program that can be solved efficiently using  off-the-shelf solvers with bounds on solution quality.
To predict the benefit of learning, we propose a precondition prediction classifier.
Given two tasks, this classifier predicts whether a skill trained on one will transfer to the other.
% use an approximate simulation model of the tasks to train a precondition model that is parameterized by the training task.
Finally, we evaluate our approach on peg insertion and Lego stacking tasks, both in simulation and real-world, showing substantial savings in human effort.

\end{abstract}

% \keywords{Shared Autonomy, Planning, Learning from Demonstrations} 

%===============================================================================

\section{Introduction}
For real world applications of robotics, like manufacturing and health-care, autonomy is not an end in itself, but a means to improve productivity and safety.
It is often infeasible or expensive to teach robots to be fully autonomous due to changing environments and task requirements.
In practice, these autonomous systems often have the option of falling back on human help when needed.
% with humans by taking over some of the tasks because it is infeasible or expensive to teach them to handle every situation that may be encountered.
% What is the optimal level of autonomy for a given set of tasks?
% in unstructured and dynamic settings.
% When teaching robots to be autonomous, it is important to 
% better characterized as semi-autonomous systems~\cite{zilberstein2015building}.
% Robots in the real world make mistakes when they come across situations they are not trained for due to changing environments and task requirements.
In this work, we consider two modes of help provided by a human-- giving demonstrations on how to do a new task and fully taking over a task.
The former allows the capabilities of a robot to be extended, which is useful when similar tasks are expected to be encountered again in the future.
The latter allows the robot to avoid attempting or having to learn one-off tasks.

Consider a manufacturing facility that gets its orders at the start of each day and needs to fulfil those orders by its end.
The factory operators may have only an approximate idea about future demand.
Hence, when the orders arrive there will be tasks that the robot can not complete autonomously.
This leads to a number of questions: Should additional robot teaching be done? If so, on which tasks? What tasks should be done by robots and what tasks by humans?

\begin{figure}
    \centering
    \includegraphics[width=1\columnwidth]{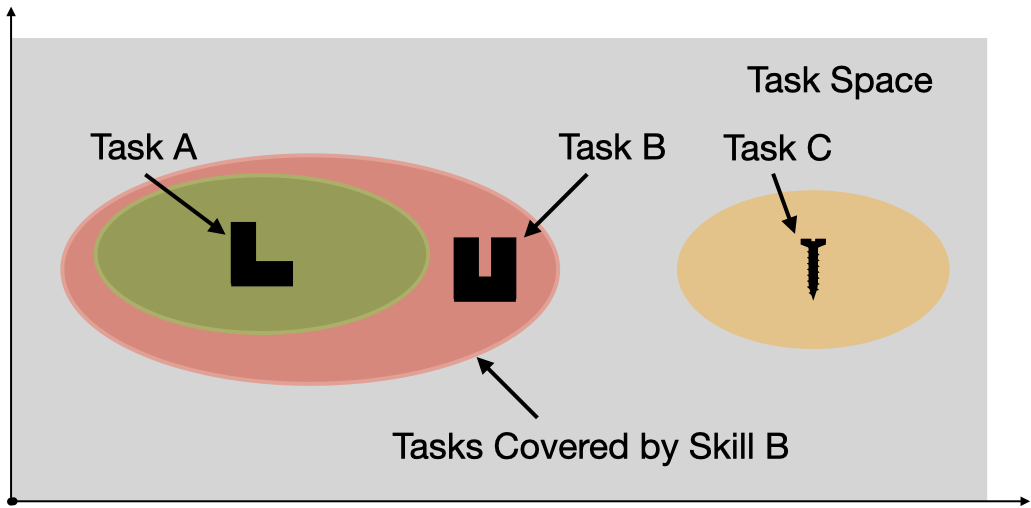}
    \caption{Consider three assembly tasks visualized in a 2D task state-space.
    Each colored oval covers tasks that can be solved by a specific robot skill.
    Note that skill B covers more tasks than the other two skills while task C remains uncovered even after learning skill B.
    Our framework schedules teaching of only those skills that cover enough future tasks to offset the cost of robot teaching.
    Remaining tasks are delegated to a human for completion.
    }
    \label{fig:motivation}
\end{figure}

To this end, we propose a decision making framework \textit{Act, Delegate, or Learn} (ADL) that jointly reasons about autonomous execution in synergy with \textit{both} of these modes of human assistance.
In particular, we look at a setting where tasks come in a fixed sequence.
This is motivated by time and cost critical domains like agile assembly lines in factories and robots in outer space, where a diverse but known set of tasks need to be accomplished with minimum human and robot effort.
While human help is available, it is at a premium.
Hence, we would like to use it optimally so as to minimize the overall effort.

Each of these two modes has been investigated individually in prior works.
A number of works~\cite{chernova2009interactive,cakmak2010designing,gribovskaya2010active,hayes2014discovering} in Learning from Demonstrations (LfD)~\cite{argall2009survey,chernova2014robot} use measures of confidence in the robot's actions and  active learning~\cite{settles2009active} to teach a robot with fewer human demos.
By not considering the option of delegating tasks to a human, these approaches seek to achieve full autonomy, which is not cost-effective in settings where delegation is possible.
In contrast, planners for task allocation~\cite{gombolay2013fast,gombolay2013towards,shannon2016adaptive} and adjusting the level of autonomy~\cite{wray2016hierarchical,basich2020learning} have been proposed for human-robot teams.
The focus on these works is on handling spatiotemporal constraints and different human models~\cite{pew1969speed,chen2014human,shannon2017human}.
However, they assume a static model of the robot's capabilities, which does not allow them to leverage robot learning in their framework.

\begin{figure}[]
    \centering
    \includegraphics[width=1\columnwidth]{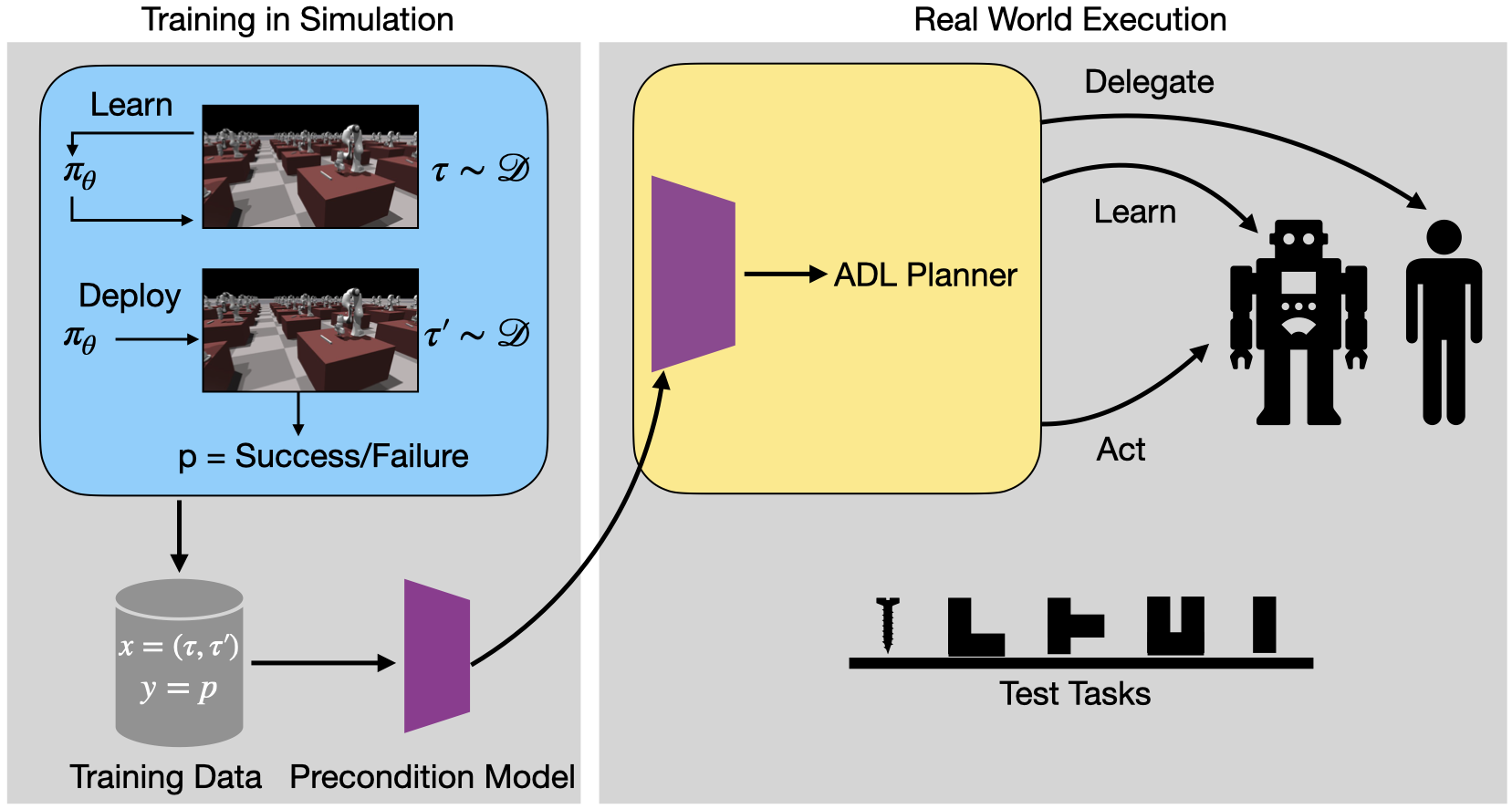}
    \caption{Overall approach: (i) \textbf{Training in Simulation} Skills are learned (using RL) and deployed on tasks $\tau \sim$ $\mathcal{D}$ to collect data on what other tasks can be solved by a skill learned for a particular task. A precondition prediction model is trained using this data.  (ii) \textbf{Real World Execution} Our planner makes use of the learned model to decide when the robot should attempt a task, when it should delegate to a human and when it should learn a new skill for a task.}
    \label{fig:method_pipeline}
\end{figure}

In summary, the main contributions of our work are:

(1) \emph{Act, Delegate or Learn Framework:} We formulate the problem of completing a given sequence of tasks with human help at minimum total expected human and robot effort as a Stochastic Shortest Path problem.
(2) \emph{Efficient Planning:} We propose a mixed integer programming formulation to efficiently solve this problem.
(3) \emph{Precondition Prediction:} Planning requires the ability to foresee the benefit of robot teaching before committing to it.
To this end, we propose a precondition prediction model that predicts what other tasks the robot will be able to solve after getting demonstrations for a task.
We train this model offline using a domain-specific simulation.
(4) \emph{Simulated and Real World Evaluation~\footnote{Videos and supplementary material are available at \url{https://sites.google.com/view/actdelegateorlearn}} :} We evaluate the benefits of our approach on two challenging manipulation tasks: (a) Peg-in-a-hole: Insert pegs into holes under uncertainty using environmental contact for localization and (b) Lego Stacking: Robustly stack complex parts made with Lego bricks onto a Lego base plate.

\section{Related Work}
\textbf{Function Allocation} is the decision making problem of determining which functions should be performed by machines and which by humans~\cite{inagaki2003adaptive,Fitts1951Human}.
While a number of strategies have been proposed, the one closest to our work is \textit{economic allocation}~\cite{inagaki2003adaptive,dearden2000allocation} which finds an allocation that ensures economic efficiency.

\textbf{Adaptive Automation} can accommodate changes in the environment or the human for function allocation.
A number of frameworks have been proposed over several decades~\cite{rouse1976adaptive,Scerbo1996TheoreticalPO,kaber2001design,sheridan2011adaptive} which focus on optimizing operator workload, attention and efficiency.
Consequently, their focus has been on modeling the human~\cite{pew1969speed,chen2014human,shannon2017human}.
\cite{basich2020learning} recently propose an interactive model of autonomy, where a system learns a model of its competence online.
All these strategies assume that the robot has certain fixed capabilities

\textbf{Learning from Demonstrations:}
There are three main categories\cite{Ravichandar2020}  of LfD-- kinesthetic teaching, teleoperation and passive observation.
Kinesthetic teaching is the most common approach for providing demos in manufacturing and health-care~\cite{Ravichandar2020}, while teleoperation does not require the user to be copresent with the robot.
Passive observation usually requires multiple demos~\cite{hayes2014discovering}, special instrumentation (motion capture, force-torque sensors) depending on the task and is complicated to solve due to the need for retargeting.
Despite recent progress, teaching robots generalizable skills still requires significant human effort.

Consequently, a number of works~\cite{chernova2009interactive,cakmak2010designing,gribovskaya2010active,hayes2014discovering} seek to minimize the number of demos required for teaching.
In particular, Confidence-Based Autonomy~\cite{chernova2009interactive} uses classification confidence to choose between autonomous execution and request for a demo.
~\cite{Rigter2020} propose an online approach to training a set of controllers from demonstrations that tries to myopically minimize human effort.
ThriftyDAgger~\cite{hoque2021thriftydagger} uses estimated probability of task success to determine when to solicit human interventions.

% Confidence-Based Autonomy (CBA)~\cite{Chernova2009InteractivePL} which uses classification confidence to choose between autonomous execution and request for a demo.
%Their method uses confidence thresholds to identify states with the greatest uncertainty and seek help when the robot is uncertain about which action to take.

\textbf{Multi-task Learning:}
~\cite{Deisenroth2014} look at learning a single policy in a multi-task setting with a continuous set of tasks.
~\cite{Kupcsik2017} learn a two level policy where the low level policy controls the robot for a given context and the high level policy generalizes among contexts.
In contrast, we take a library of independent skills approach, where generalization happens only at the lower level.

\section{Preliminaries}
\textbf{Skill Preconditions:}
We model skills using the options framework~\cite{sutton1999between,konidaris2009efficient,kroemer2019review}.
% The preconditions~\cite{konidaris2015symbol} of a skill refer to the circumstances in which it can be executed to achieve a desired goal.
We use a probabilistic notion of skill preconditions~\cite{konidaris2015symbol}, where the preconditions of a skill is a classifier $\rho: \Theta \rightarrow [0, 1] $ that takes in features describing a task and returns the probability that the skill will be able to successfully complete the task.
This classifier is usually trained by executing the skill on a distribution of tasks to generate success/failure labels~\cite{kroemer2016learning,sharma2020relational}.
However, this is an expensive process which  requires real world execution of the robot.

\textbf{Skill Library:}
A popular approach for solving related tasks is to learn a parameterized skill~\cite{konidaris2012learning}, that adapts the policy based on changes in the task.
This approach is practical if only some aspects of the task can change.
Adapting to various changes in the tasks requires a more complex skill parameterization that makes the learning problem harder and more sample complex.
An alternative approach, which we take in this work, is to have the robot maintain a library $\mathcal{L} = \{\pi_1,\cdots,\pi_n\}$ of skills, each of which is learned on a narrow task distribution from demonstrations.
Given a task $\tau$, the robot picks an appropriate skill for it.
by selecting a skill with the highest probability of success: $\argmax_{\pi \in \mathcal{L}} \rho_{\pi}(\tau)$.
This representation has a number of advantages over learning a monolithic skill, chiefly, modularity, allowing local updates and providing alternatives in case of execution failure.

%\subsection{Weighted Set Cover Problem}
%~\citet{vazirani2013approximation} give the following definition of the set cover problem:

%\textit{Given a universe $U$ of $n$ elements, a collection of subsets of $U$, $\mathcal{S} = \{S_1,\cdots,S_k\}$, and a cost function $c: \mathcal{S} \rightarrow \mathbb{Q}^+$, find a minimum cost suballocation of $\mathcal{S}$ that covers all elements of $U$.}

\section{The Act, Delegate or Learn Framework}
\begin{figure*}[h!]
    \centering
    \subfloat[][]{
        \includegraphics[width=0.56\textwidth]{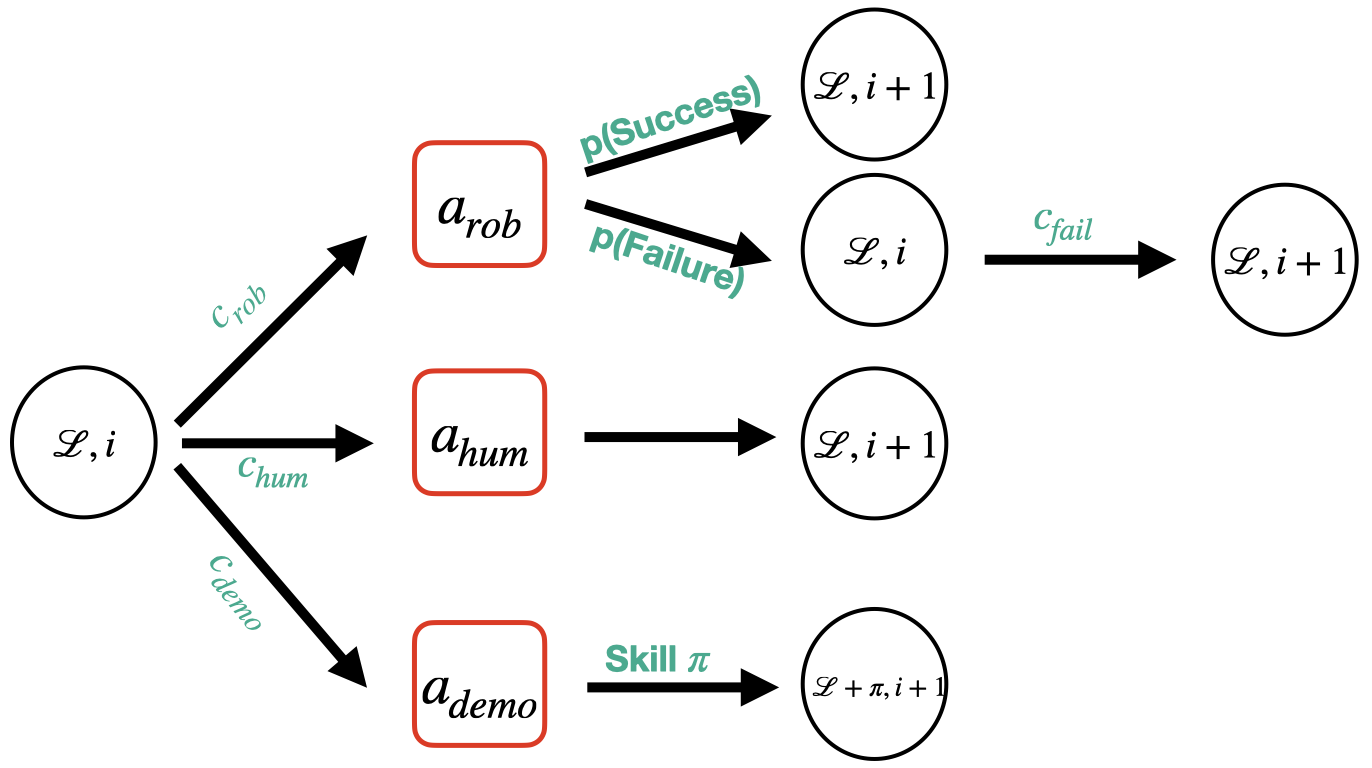}
    }
    \subfloat[][]{
        \includegraphics[width=0.41\textwidth]{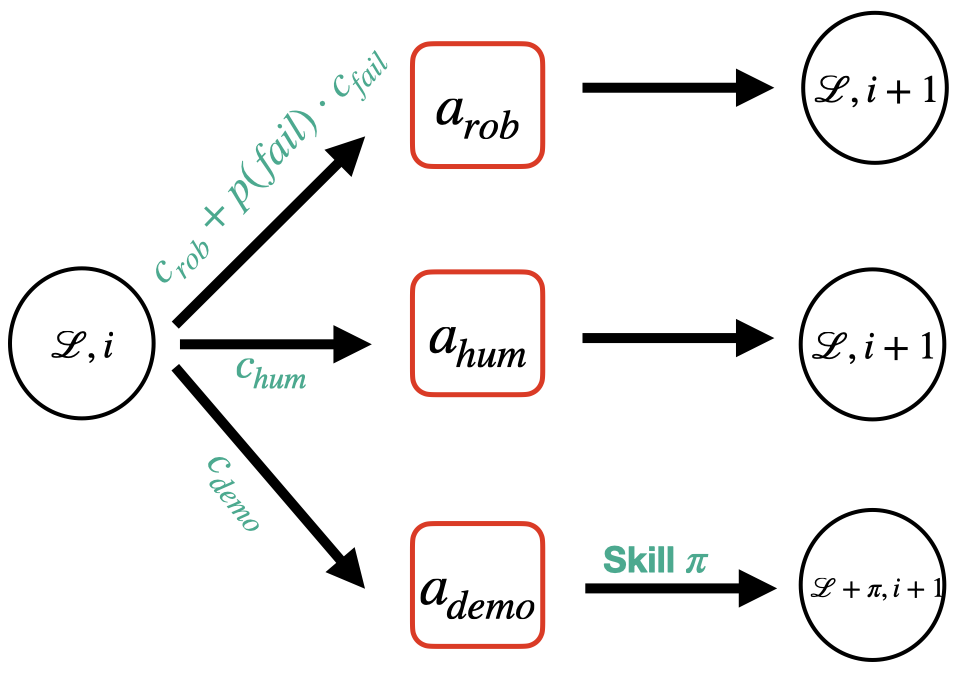}
    }
    \caption{(a) \textbf{Transition Model:} Our MDP has three actions: $a_{rob}$, $a_{hum}$ and $a_{demo}$ with associated costs of $c_{rob},$ $c_{hum}$ and $c_{demo}$ corresponding to the options act, delegate and learn. A human intervenes to complete a task if robot execution fails. We assume that a human can complete all the tasks and is available at all times to teach the robot. (b) \textbf{Simplified Transition Model:} We can replace the two stochastic outcomes due to $a_{rob}$ with a single outcome whose cost is an expectation over them.}
    \label{fig:mdp}
\end{figure*}

We are interested in completing a sequence of tasks with minimum total expected human and robot effort.
At train time, we are provided a distribution $\mathcal{D}$ of tasks that are expected to be encountered.
The robot may be pre-trained with a set of skills based on this knowledge.
The actual tasks and the order in which they need to be done are revealed only at test time.
In this stage, a decision needs to be made for every task: should the robot do the task, should it delegate the task to a human or should it ask to be taught how to do the task?
We require that every task be completed.
Hence, each robot failure incurs additional cost due to human intervention to complete the task and correct the setup.
Finally, we assume that a human is available at all times to intervene if needed - either to correct a robot failure or to teach it, for example, by providing demonstrations.

\subsection{Problem Formulation}
We formulate this problem as a Stochastic Shortest Path (SSP) problem  $(\mathcal{S}, \mathcal{A}, \mathcal{T}, \mathcal{C}, \mathcal{G})$~\cite{kolobov2012planning} where $\mathcal{S}$ is a state space, $\mathcal{A}$ is an action space, $\mathcal{T}: \mathcal{S} \times \mathcal{A} \times \mathcal{S} \rightarrow [0, 1]$ is a transition model, $\mathcal{C}: \mathcal{S} \times \mathcal{A} \times \mathcal{S} \rightarrow \mathbb{R}^+$ is a cost function and $\mathcal{G} \subset \mathcal{S}$ is a set of goal states.
We define each of these components of the MDP for our problem:

\textbf{State Space}: Each state $s \in \mathcal{S}$ is a tuple $\langle \mathcal{L}, k \rangle$, where $\mathcal{L}$ is the skill library of the robot at that state and $k$ refers to the tasks completed so far.

\textbf{Action Space:} $\mathcal{A} = \{a_{rob}, a_{hum}, a_{demo}\}$, where $a_{rob}$ implies that the robot attempts to solve the task, $a_{hum}$ implies that the human solves it and $a_{demo}$ implies that the human teaches the robot a new skill for it in addition to solving it.
% We compare these actions in table~\ref{table:actions}.

\textbf{Transition Function} $\mathcal{T}$ models whether the skill library got updated or a task was completed after an action.
Though the outcome of robot execution is stochastic, we can convert it into a deterministic MDP by taking an expectation over the two outcomes (see figure~\ref{fig:mdp} for details).
We will be using the resulting simplified transition model in the rest of the paper.
%\mathcal{S} \times \mathcal{A} \rightarrow \mathcal{S}$ as follows (see figure \ref{fig:mdp}):
    %$$\mathcal{T}(\langle P, i \rangle , a) = \begin{cases}
    %\langle P, i + 1 \rangle &  a \in \{a_{rob}, a_{hum}\} \\
    %\langle P', i + 1 \rangle  & a = a_{demo}
    %\end{cases}
    %$$
The transition model makes it clear that $a_{rob}$ and $a_{hum}$ do not affect the skill library in any way.
On the other hand $a_{demo}$ updates the library by adding a new skill $\pi$ to its repertoire.

\textbf{Cost Function}
The cost function is defined as
$$
\mathcal{C}(s_i, a) = \begin{cases}
    c_{rob}(i) +  \Pr(fail)\cdot c_{fail}(i) & a = a_{rob} \\
 c_{hum}(i) & a = a_{hum} \\
c_{demo}(i) & a = a_{demo} \\
\end{cases}
$$
where, $\Pr(fail) = 1 - \max_{\pi \in \mathcal{L}}\rho_{\pi}(\tau_i)$. 
The cost of a robot execution includes the cost of a potential failure and hence depends on the robot's skill library.
$c_{rob}$, $c_{hum}$ and $c_{demo}$ are domain and task dependent costs specified by a domain expert.
For example, in manufacturing, where minimizing the \emph{economic cost} of production is crucial, $c_{rob}$ could reflect the cost of operating a robot, while $c_{hum}$ and $c_{demo}$ could depend on the efficiency of a human collaborator.
There exist a number of approaches~\cite{chen2014human,shannon2017human,gombolay2017computational} to model human performance.
% $c_{hum}$ could reflect human time and $c_{demo}$ could the difficulty of teaching the robot.
$c_{fail}$ corresponds to the difficulty of fixing a mistake made by the robot.
In some domains, this could be as simple as asking a human in the factory to complete the remaining task, while in others, it may be high if there is a risk of damage due to a failure.

\textbf{Goal:} A goal state is reached once all the tasks have been completed.

Let $\{\tau_i\}_{i=1}^n$ be the sequence of tasks and $\eta = \{\eta_i\}_{i=1}^n$ be the sequence of actions taken.
Then, the expected cost of execution is: $J(\eta) = \sum_{i=1}^{n}\mathcal{C}(s_i, \eta_i)$, where $s_{i+1} = \mathcal{T}(s_i, \eta_i)$ and our goal is to find an optimal plan $\eta^* = \argmin_{\eta \in \mathcal{A}^n}{J(\eta)}$.

%\begin{equation}\label{eq:objective}
  %\end{equation}
%Note that the robot's probability of success is not stationary but changes over time as it is taught new skills.

% We can discard two degenerate cases right away:
% \begin{enumerate}
%     \item If $a_{hum}$ is the cheapest action, then the optimal plan is that the human does all the tasks.
%     \item If $a_{demo}$ is the cheapest action then the human can teach the robot how to do all the given tasks.
% \end{enumerate}

% We would like the robot to solve most of the tasks, but it cannot do so until it has been taught the relevant skills. On the other hand, it takes much more effort by a human to teach a new skill to the robot than to do the task himself. This presents a dilemma - for every task, should the human do it himself or should he teach the robot how to do it so that it can solve similar tasks in the future?

\section{Planning}
% Is there a simple optimal strategy for this problem?
% We answer this question in the negative in the following section by showing that the problem is at least NP-hard.
% This necessitates planning to solve the SSP.
The standard techniques used to solve a deterministic SSP are graph search algorithms like Dijkstra's algorithm and A*.
Unfortunately, the search graph induced by our problem has \textit{exponentially} many states in the number of tasks to be done.
Though A*-like algorithms can leverage heuristics to speed-up search, their performance is highly dependent on the quality of the heuristic and hence incur substantial overhead for designing good heuristics.

Motivated by this, we propose a mixed integer programming (MIP) formulation of the SSP which can be solved using off-the-shelf solvers without the need to design heuristics.
These solvers provide high quality solutions (with sub-optimality bounds) and are highly scalable.

\subsection{MIP Formulation}
%lego_robot_action.png\begin{definition}
%We say that a skill \textbf{covers} a task if it can solve it.
%The \textbf{coverage} of a skill is the set of tasks it can solve.
%\end{definition}
We introduce decision variables for every task: $x_i, y_i, z_i \in \{0, 1\}, w_i \in [0, 1], \forall i \in \{1,\cdots,n\}$.
Let binary decision variable $x_{i}$ be 1 if a demo is sought on task $\tau_i$, ${y_i}$ be 1 if a human is asked to solve it and $z_{i}$ be 1 if the robot is asked to attempt the task.
As the robot may fail in its attempt, we model the probability of human intervention with a continuous decision variable $w_i$-- note that it is non-zero only if robot execution is chosen for a task.
We exercise indirect control over $w_i$ via the probability of failure of the action taken.

Our overall objective is:
    $$\min \sum_{i=1}^{n}{c_{demo}(i) x_{i} + c_{hum}(i) y_i + c_{rob}(i) z_i + c_{fail}(i)  w_i}$$
% \end{align}

where, $z_i = 1 - x_i - y_i$ as we allow exactly one of these three actions for a task. 
Hence, the objective can be simplified.
\begin{align}\min \sum_{i=1}^{n}{c_{demo}'(i)x_{i} + c_{hum}'(i) y_i + c_{fail}(i) w_i}\end{align}
where $\forall i \in \{1,\cdots,n\}$
\begin{align}
    &c_{demo}'(i) = c_{demo}(i) - c_{rob}(i )\nonumber \\ 
    &c_{hum}'(i) = c_{hum}(i) - c_{rob}(i) \nonumber \\ 
    &w_i = 1 - \max{\{\rho_0, \rho_1(\tau_i)x_1, \cdots, \rho_i(\tau_i)x_i, y_i\}} \nonumber
\end{align}
The max term in the last equation is a maximization over the success probabilities of the available ways to solve the task -- using pre-trained skills (with precondition $\rho_0$), learning new skills (with preconditions $\rho_1\cdots,\rho_n$) and delegating to a human (represented by $y_i$).
$y_i$ is 1 if the robot delegates the task to a human, in which case we are assured of task completion.
In its current form this program is not linear due to the $\max$ operation.
However, we can easily convert it into a linear MIP by introducing additional binary decision variables.
Some solvers like Gurobi~\cite{gurobi} can directly take this program and do the linearization under the hood.

\textbf{Note:} Alternatively, we can also formulate it as a facility location problem~\cite{vazirani2013approximation}, where all tasks are customers and opening a facility corresponds to either seeking a demo or delegating to a human.
While solving the facility location problem optimally is NP-hard, it has $O(\log n)$ approximation algorithms which can be useful if the MIP is too big to solve optimally.

\section{Precondition Prediction Model}
\begin{figure}[t]
\centering
% \subfloat[]{
    \includegraphics[width=0.9\columnwidth]{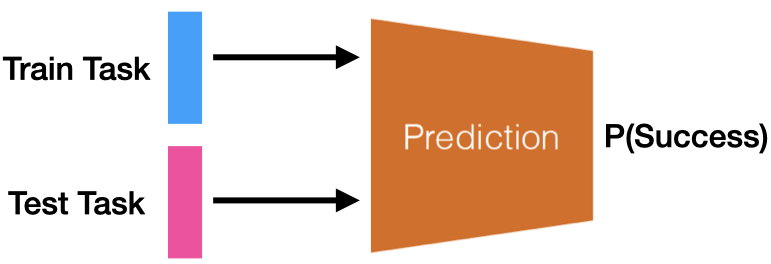} 
    % }
% \subfloat[]{
%     \includegraphics[width=0.9\columnwidth]{abstract_sim.png} \label{fig:abstract_sim}
%     }
    \caption{Precondition prediction model predicts the probability of success on a test task $\tau'$ after the robot has been trained on a task $\tau$.}
    % \emph{(b)} Abstraction M simplifies tasks such that inter-task relationships remain intact.}
    \label{fig:precond_net}
\end{figure}

A key requirement of our planner is the ability to foresee the benefit of robot teaching \emph{before} committing to it.
Past works~\cite{kroemer2016learning,sharma2020relational} have looked at the problem of precondition learning, wherein a classifier is trained for an existing skill to predict what other tasks can be solved by it.
By contrast, we need to predict the preconditions of a skill that \emph{will} be learned if we choose to teach the robot-- a precondition prediction problem.
Our proposed solution is to learn a classifier (see figure~\ref{fig:precond_net}) that takes as input a train task and a test task and predicts whether a robot trained on the former will be able to solve the latter.
Intuitively, this can be thought of as learning a similarity metric between tasks.
% Our proposed solution is to train a task-parameterized precondition model on a range of tasks so that it can generalize to unseen tasks.

We collect training data for the precondition model using algorithm~\ref{alg:data-generation}.
This can be prohibitively expensive as we need to learn robot policies to generate labels.
% ground truth data in real world or in a high-fidelity simulation to train this model can be prohibitively expensive. 
We get around this limitation by observing that we do not need to transfer robot skills from sim2real but only task relationships--- the former requires high fidelity simulation while the latter does not.
It is often the case that a lower dimensional state representation is sufficient to discriminate between tasks.
The key is to simplify the problem such that inter-task relationships remain intact-- tasks that are similar/dissimilar in the real world should remain so in simulation and vice versa.
Concretely, we define an \emph{abstraction}~\cite{li2006towards,konidaris2009efficient} $M$ as  a pair of functions $(f, g)$ such that $f: S \rightarrow S'$ maps the original problem state space $S$ to a smaller state space $S'$ and $g: A \rightarrow A'$ maps the full action space to a smaller action space.
The specific state and action abstraction to be used in training are provided as domain knowledge.

\begin{algorithm}
\centering
    \begin{algorithmic}[1]
        \Procedure{GetTrainingData}{$m, n$}
            \State{\textsc{X} $\leftarrow \phi$, \textsc{Y} $\leftarrow \phi$}
            \State{\textsc{S} $\leftarrow$ \textsc{Sample} $m$ \textsc{tasks from} $\mathcal{D}$} 
            \For{$i \in \{1,\cdots,n\}$}
                \State{\textsc{Sample $\tau$ from $\mathcal{D}$}}
                % \State{$\tau \sim f_X(x)$}
                \State{\textsc{$\pi \leftarrow$ Learn policy for $\tau$}}
                \For{$\tau'$ $\in$ \textsc{S}}
                    \State{x $\leftarrow (\tau, \tau')$}
                    \State{y $\leftarrow$ \textsc{Evaluate} $\pi$ \textsc{on} $\tau'$}
                    \State{\textsc{X.insert(}x), \textsc{Y.insert(}y)}
                \EndFor
            \EndFor
            \Return{X, Y}
        \EndProcedure
    \end{algorithmic}
    \captionof{algorithm}{Data collection in abstract simulation.}
    \label{alg:data-generation}
\end{algorithm}

\section{Experiments}

We evaluate our approach, both in simulation and in the real world, on two challenging problems (1) block insertion under uncertainty and (2) Lego stacking.
Our objectives are (1) to understand the benefits of the ADL framework as compared to baselines that are myopic or reason about only a subset of the three options
(2) to evaluate our hypothesis that the precondition model can be trained in simulation.
In both these experiments, we use a 2-layer fully connected neural network as our precondition prediction model and we are able to solve our mixed integer program optimally in well under a second using Gurobi~\cite{gurobi}.
We provide additional details in the appendix.

\begin{figure}[t]
    \centering
    \includegraphics[align,width=1\columnwidth]{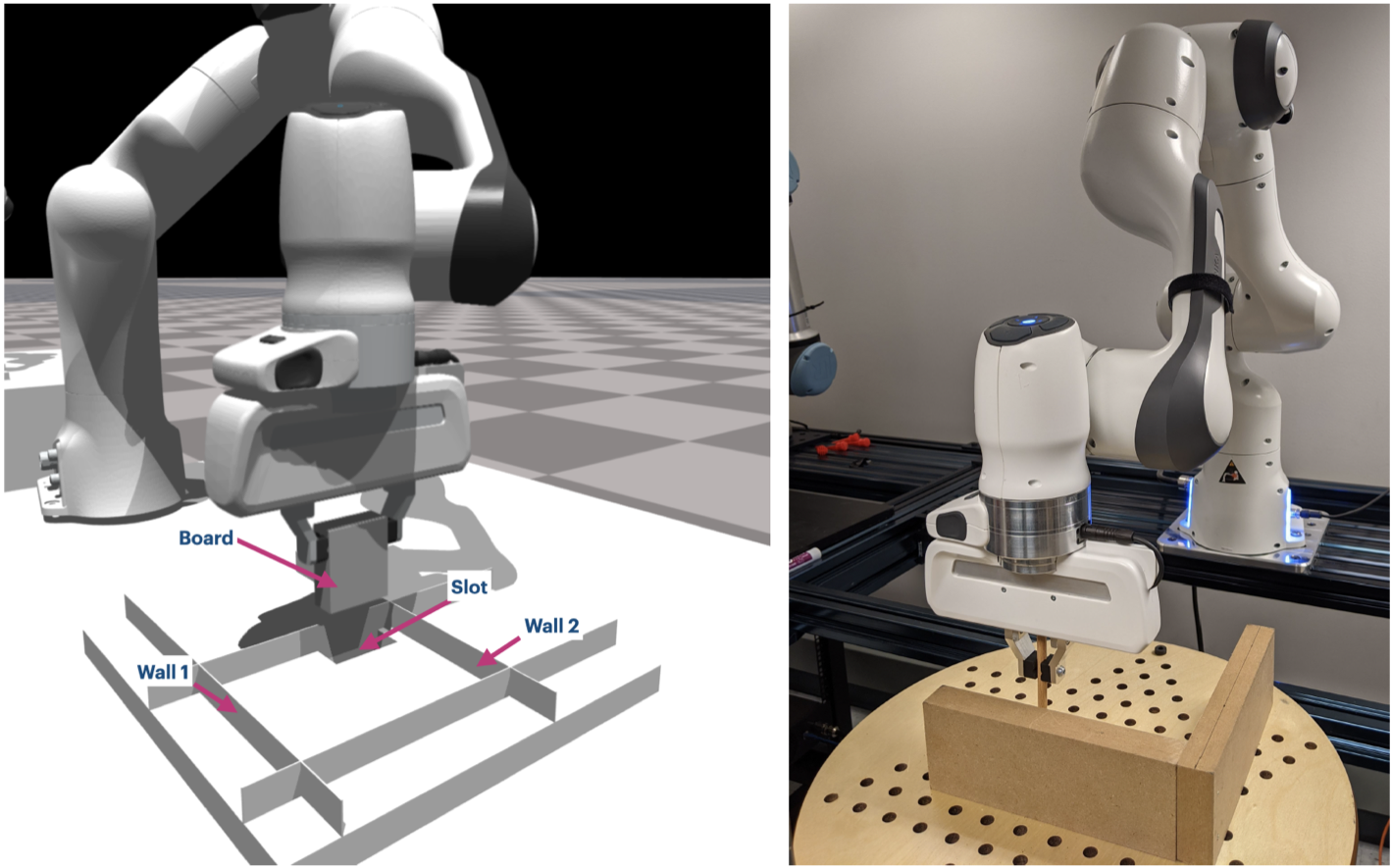}
    \caption{Block (peg) insertion under uncertainty in simulation and in real world.}
\end{figure}

\textbf{Baselines:} We compare our approach against three baselines: 
(1) \emph{Act Delegate (AD):} The robot chooses between acting and delegating based on the expected costs of these two actions.
(2) \emph{Confidence-Based Autonomy (CBA)~\cite{chernova2009interactive}:} Given a fixed threshold $\theta$, the robot attempts a task if its confidence in success is greater than $\theta$ and asks for demonstrations, otherwise.
(3) \emph{Act-Learn Myopic (ALM):} Similar to the strategy used by \cite{Rigter2020}, the robot chooses between attempting a task and asking for human demos by comparing the immediate expected costs of both the actions.

\textbf{Metrics:} The main evaluation metric is the total cost of completing all the given tasks.
We also compare the methods based on the number of demonstrations and human interventions and the number of failures.

\textbf{Skill Representation:} In both our experiments, the robot end-effector is controlled using Cartesian-space impedance control which commands torques at the end-effector based on errors in the Cartesian space using a spring-damper system.
A skill is a sequence of waypoints in the robot’s end-effector frame, where each waypoint is defined by a 6D pose and the stiffness to be used in the corresponding spring-damper system.
% For safety reasons, we also set a max-force limit at the end-effector.

\subsection{Block Insertion}
Our first evaluation is in simulation to understand how well our planner performs in comparison with standard non-planning approaches.

\textbf{Task:} Each task involves inserting a block of dimensions 1 cm x 1 cm x 6 cm into a slot of dimensions 1.2 cm x 1.2 cm x 2 cm in a known environment with a noisy estimate of the slot location $\sim \mathcal{N}(0,0.3^2 cm^2)$.
% We compute a 4D feature vector for each task by tracing rays starting from the center of the slot in four axis-aligned directions.
We generate four different environments of dimensions 20 cm x 20 cm each, with different numbers of walls arranged in a grid.
We use the  Nvidia Isaac Gym  simulator~\cite{ig2020} to simulate the tasks and to train the precondition model.
% to learn and evaluate robot policies on the Franka Emika Panda arm.

% \textbf{Skill} Each skill is a sequence of 4 robot waypoints in the end-effector frame that are passed on to a Cartesian-space impedance controller.

\begin{figure}[t]
    \centering
        \includegraphics[align=b,width=1\columnwidth]{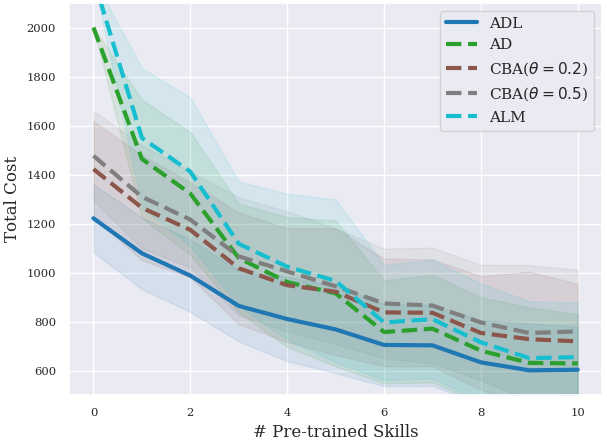}
    \caption{Comparison of ADL vs baselines in total cost for solving 20 block insertion tasks at different levels of skill pretraining.  Pretraining is done by teaching the robot randomly sampled tasks from the task distribution. ADL is strictly better than all baselines at every level of pre-training. However, after pre-training with 8 skills, both ADL and AD converge to full autonomy as the robot is able to solve most of the tasks with pre-trained skills. We use $c_{rob}=10$, $c_{hum} = c_{fail} = 100$ and $c_{demo} = 200$.}
    \label{fig:block_insertion_cost_plot}
\end{figure}

\textbf{Simulation Results:} 
We compare ADL with AD, CBA($\theta=0.5)$ , CBA($\theta=0.2)$ and ALM in figure~\ref{fig:block_insertion_cost_plot}, where $0.2$ is the optimal CBA threshold found using grid-search.
ADL outperforms all the baselines at every level of pretraining.
However, the improvement provided by ADL drops with increase in pretraining as the robot can complete more of the tasks autonomously without seeking additional demos or delegating.
Also note that CBA outperforms AD at low levels of pretraining but the opposite holds at higher levels as demos sought by CBA are not cost-effective for the task set.
We provide comparisons using different costs and qualitative results from a real-world experiment in the appendix.

\subsection{Lego Stacking}
In our second domain of Lego stacking we seek to evaluate how well our method works in the real world.
In particular, we want to understand whether a precondition model learned using an abstract simulation is able to reduce effort in real world.
% Most physics-based simulators struggle with interaction of objects at close distances and high forces.
% The Lego interlocking mechanism is an interference fit and parts consisting of Lego bricks are not rigid--- these reasons make it hard to learn learn policies in simulation.
% Hence, human demonstrations are the most practical way to teach robots in this case.

% Despite being a toy, it is challenging to teach Lego stacking to a robot with just trial and error due to complex interactions between Lego bricks and studs.
% Prior works~\cite{Popov2017DataefficientDR,haarnoja2018composable} have used RL to learn stacking policies for Lego Duplo\footnote{A Lego Duplo brick is twice the size of a standard Lego brick along every dimension.} bricks but not for standard Lego bricks.
% Hence, human demonstrations and guidance is critical in this domain.
% Here, we teach the robot to robustly stack a variety of complex parts different stacking strategies depending on their geometry.

\begin{figure}[!h]
    \centering
    % \subfloat[][]{
        \includegraphics[align=b,width=0.95\columnwidth]{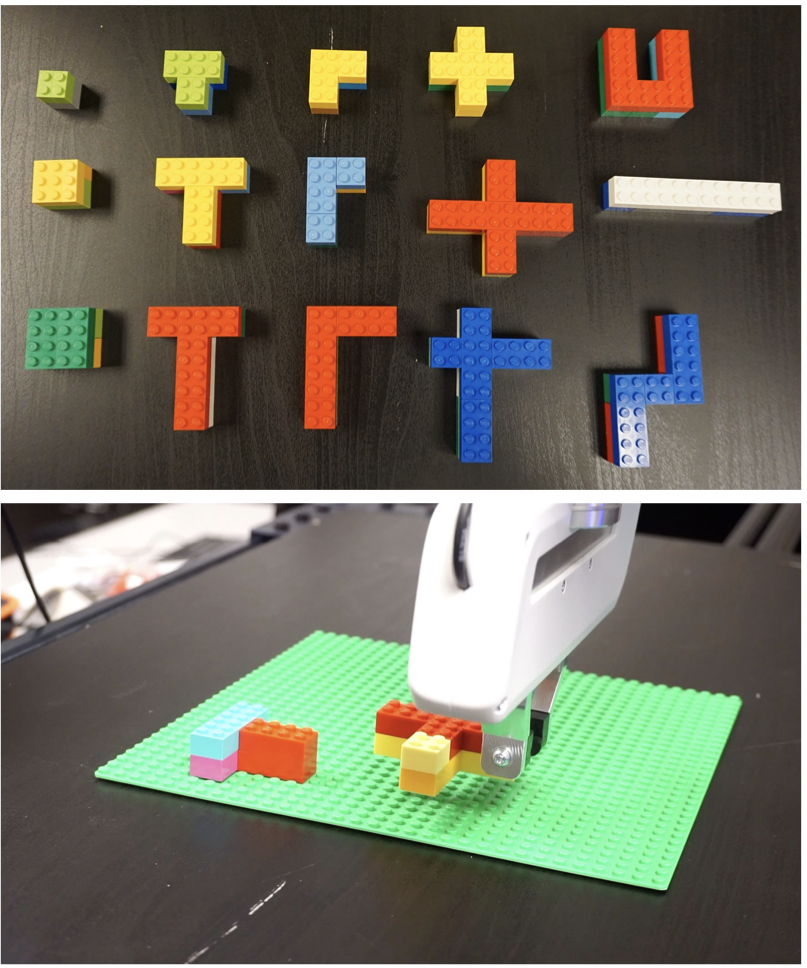}
    % }
    
    % \subfloat[][]{
        % \includegraphics[align=b,width=0.7\columnwidth]{data/lego_robot_action.png}
        \label{fig:preconds}
    % }
    \centering
    \caption{(\emph{Top}) The ground set of 15 tasks from which test sets of 10 tasks each are sampled uniformly randomly. (\emph{Bottom}) The Franka-Emika Panda robot stacking one of the parts onto the base plate.}
    \label{fig:lego_tasks}
\end{figure}

\textbf{Task:} Each task involves picking up a part made up of Lego bricks from a table and stacking it firmly onto a Lego base plate.
A robot execution fails if two or more corners of the part are not locked onto the plate or the robot hits the base at any point.
The robot is provided a bounding box around the part, a grasp location and a target location by the user.
We use a 66D feature vector for each task-- binarized and resized image (to $8 \times 8$ ) along with its original size.
Before running the experiments, we record 5 demos for each task in the ground set.
Every time the robot requests a demo for a task, one of the 5 pre-recorded demos is provided by sampling randomly.

\textbf{Skill:} Each stacking skill consists of three sub-skills executed in sequence: pickup, place-and-wiggle and robust-tapping.
The first two are hand-designed and common across all tasks, while robust-tapping needs to adapt the number and location of taps based on the geometry of the part.
The latter is learned in the grasp-frame and scaled based on the size of the part.
This allows the skill to generalize to different locations and across parts of similar shape but different sizes.

\textbf{Data Collection:}
Physics-based simulators struggle to simulate interactions among multiple Lego bricks and the interference fit mechanism used in them. 
% while the pieces used in constructing parts consisting of Lego bricks are not rigid.
% This made make it hard to learn learn policies in simulation.
% Hence, human demonstrations are the most prac
Consequently, we use a custom simulation based on our observation that the primary reason for variability in skills is the geometry of the parts.
We can afford to ignore physics and robot dynamics as we do not transfer the learned skills to the real world.
Our coverage-based simulation takes in a 2D image of a part and identifies only the number and location of taps needed to cover the whole part by randomly sampling points on the image.
Experimentally, we found that a single tapping action has an effect  upto about 3cm from the tapping location.
We use this knowledge in the simulation to determine whether a part is covered or not after a sequence of taps.
% As we can see in figure~\ref{fig:preconds}, inter-relationship between parts in this abstract model approximately matches the inter-relationship in the real world.
We capture 10 images  of each of the 15 tasks, along with a bounding box around the part and the grasp location.
After training skills for each of the resulting 150 tasks in our coverage-based simulation, we evaluate them on all the tasks to generate binary success labels.

\begin{figure}[t]
\centering
\includegraphics[width=0.95\columnwidth]{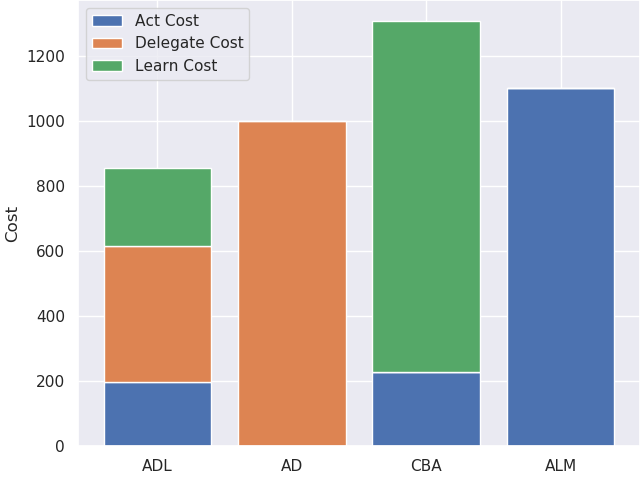}    
\caption{Comparison of ADL vs baselines in the Lego stacking domain using $c_{rob} = 10, c_{hum} = c_{fail}= 100$ and $c_{demo} = 200$. ADL is the only planner that leverages synergy among acting, delegating and learning to complete tasks at minimum cost.
}
\label{fig:lego_bar_plot}
\end{figure}

\textbf{Real World Results:} We evaluate all the approaches on 10 sets of 10 Lego-stacking tasks each
using $c_{rob} = 10, c_{hum} = c_{fail}= 100$ and $c_{demo} = 200$.
We choose $c_{demo} > c_{hum}$ as it takes much more time to provide a demo than for the human to stack the Lego themself, while $c_{hum} = c_{fail}$ as a failed robot execution can be fixed quickly by a human.
$c_{rob}$ is the smallest cost as we value human time much more than robot time in this domain.
Figure~\ref{fig:lego_bar_plot} shows the total cost of completing all tasks using each of the methods.
AD delegates all tasks as the skill library is empty at the beginning, CBA asks for too many human demos as it doesn't take into account their relevance to the task set and ALM doesn't ask for any demos as its upfront cost is higher than failing at a task.
In contrast, ADL finds the optimal synergy among all the three options to solve the tasks with minimum cost.
% The full results are summarized in table~\ref{table:results1} in the appendix.
% This also validates our hypothesis that a parameterized precondition model trained in an abstract simulation can be useful in the real world.

% \subsection{Effect of Costs on Plans}

\section{Conclusion and Future Work}
 We propose a planning and learning framework for completing $n$ tasks with a human-robot team using minimum total effort.
Our approach has two key components: (1) a \emph{general} mixed integer programming formulation and (2) a learned \emph{ domain-dependent} precondition prediction model to predict the benefits of learning a new skill.
Simulated and real world evaluations on two challenging manipulation domains indicate that our approach saves significant human and robot effort compared with approaches that do not plan ahead.

In the future, we are interested in extending the planner so that it can also optimize the order of tasks.
We would also like to continue working on the precondition prediction problem to make it less data-hungry and more accurate by using multiple sources of data.
% further explore sources of data to learn the precondition prediction problem to make it less reliant on only simulation data.
% For example, we observed in the Lego domains that the learned model makes certain mistakes because it is unaware of the gripper.
% In some domains it may be non-trivial to come up with an effective abstraction.
% In future, we plan to address these limitations by exploring  alternative sources of information like task or policy similarity metrics, alternative models for precondition prediction, and ways to use sparse real-world data to efficiently patch the precondition model.
Finally, a major limitation of the precondition prediction model is that it currently assumes each skill is trained on only one task.
We would like to extend this to skills that are trained on a set of tasks which will allow the use of parameterized robot skills in our framework.

\section{Acknowledgement}
The authors thank Kevin Zhang for help with robot experiments and Jayanth Krishna Mogali for discussions.
This work is supported by ONR Grant No. N00014-18-1-2775 and ARL grant W911NF-18-2-0218 as part of the A2I2 program.


\clearpage
\balance
\bibliographystyle{IEEEtran}
\bibliography{IEEEabrv,local_bibfile,master_bibfile}
% \newpage
% \input{includes/9_appendix}

\end{document}